\documentclass[runningheads]{llncs}
\usepackage{graphicx}
\usepackage{xcolor}

\usepackage{booktabs}    
\newcommand{\tabincell}[2]{\begin{tabular}{@{}#1@{}}#2\end{tabular}}
\usepackage{caption}    
\usepackage{subfigure}
\usepackage[colorlinks=true, linkcolor=red, anchorcolor=blue,citecolor=green]{hyperref}

\usepackage{geometry}
\begin{document}
\title{PyramNet: Point Cloud Pyramid Attention Network and Graph Embedding Module for Classification and Segmentation}
%
%
\author{Kang Zhiheng\inst{1}, Li Ning\inst{1}}
%
%
\institute{Department of Automation, Shanghai Jiao Tong University, Shanghai, China 
\email{\{kangzhiheng,ning\_li\}@sjtu.edu.cn}
}
\maketitle              
\begin{abstract}
Point cloud is an important 3D data structure, which can accurately and directly reflect the real world. In this paper, we propose a simple and effective network, which is named PyramNet, suitable for point cloud object classification and semantic segmentation in 3D scene. We design two new operators: Graph Embedding Module(GEM) and Pyramid Attention Network(PAN). Specifically, GEM projects point cloud onto the graph and practices the covariance matrix to explore the relationship between points, so as to improve the local feature expression ability of the model. PAN assigns some strong semantic features to each point to retain fine geometric features as much as possible. Furthermore, we provide extensive evaluation and analysis for the effectiveness of PyramNet. Empirically, we evaluate our model on ModelNet40, ShapeNet and S3DIS.

\end{abstract}
\section{Introduction}

With the advent of new concepts such as autopilots, high-precision maps and smart cities, many scenarios require perception of 3D environment perception and interaction based on point cloud. The rapid development of 3D scanning technology makes the acquisition of point cloud more simple and convenient. Hence, point cloud has gradually become a popular 3D data expression in the field of deep learning on 3D data. Point cloud data can be applied to classification, semantic segmentation, 3D object detection, 3D reconstruction \cite{3D_reconstruction_1,3D_reconstruction_2}, registration ~\cite{3D_Registration_1}, retrieval and other point cloud challenging tasks.

Nevertheless, unlike image, point cloud is highly sparse and unordered in space due to uneven sampling, sensor accuracy or other factors. PointNet~\cite{PointNet} attempts to solve these problems using symmetric functions and multi layer perceptrons, which is the pioneering application of CNN to directly consume raw point cloud. Many methods\cite{DGCNN,PointNet++,PvNet,RGCNN,Voxel_1} based on PointNet have achieved good performance 

\begin{figure}
	\includegraphics[width=\textwidth]{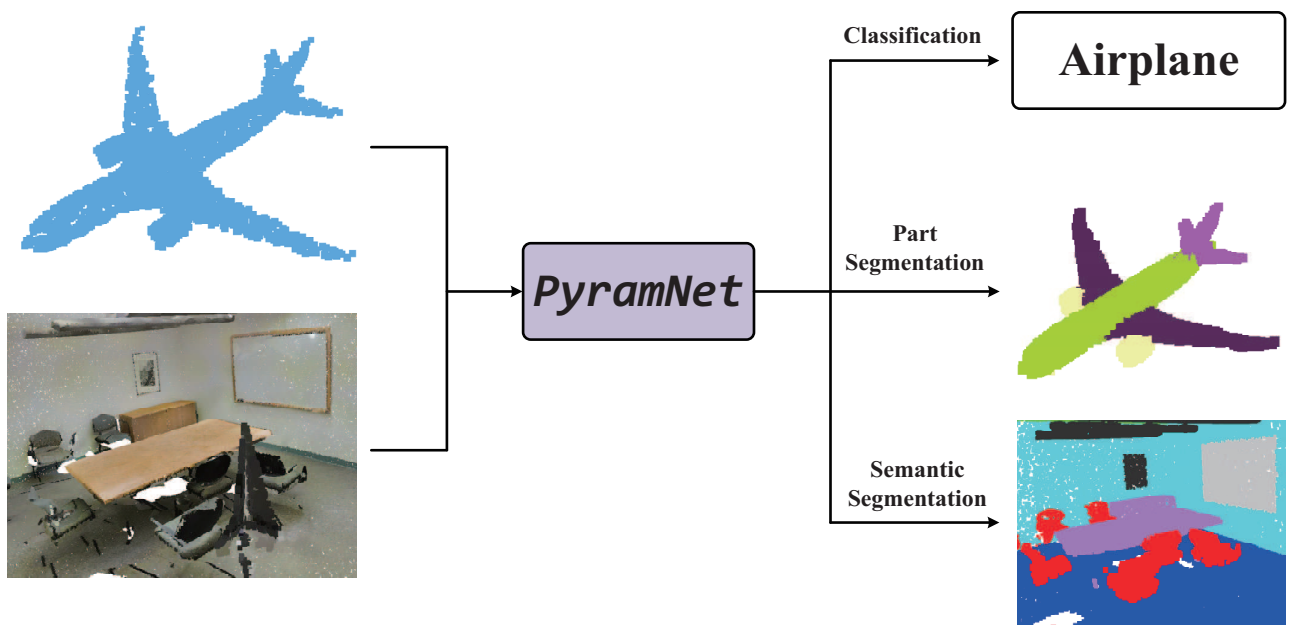}
	\caption{\textbf{3D point cloud challenge tasks}. We propose a novel deep learning architecture on point cloud to perform classification, part segmentation and semantic segmentation.} 
	\label{fig:3Dtasks}
\end{figure}

EdgeConv~\cite{DGCNN} attempts to discover the geometric relationship of point cloud. EdgeConv extracts the edge feature by the relationship between the central point and the neighbor points. However, the local feature extraction in EdgeConv is based on the Euclidean distance. Calculating the Euclidean distance in high-dimensional space does not only consume a lot of memory, but also has little practical significance. When the feature is mapped directly to high-dimensional space, the geometric relationship of point cloud may be lost.

Graph Convolutional Neural Network (GCNN) is a neural network that extends CNN to graphs. The GCNN is well suited to process irregular data structures like point cloud. We propose a novel operator, which is named \textit{Graph Embedding Module}(GEM), for extracting local geometric relationship in point cloud. We associate point cloud with the graph and use the similarity between points as the basis for feature extraction.

Most of the PointNet-based methods use multi layer perceptrons to transfer features. The only change is the number of channels per feature map, so as to enrich the semantic features of each point. The details of the geometric feature tend to get lost. Also, most points may be assimilated. To preserve the geometric details of each point as much as possible, we propose another new operator, which is named \textit{Pyramid Attention Network}(PAN). The PAN can assign some strong semantic features of each point while increasing the receptive field.

Combined with the characteristics of the Pyramid Attention Network and the Graph Embedding Module, we propose a novel end-to-end network structure named PyramNet(See Fig. \ref{fig:PyramNet}), that can consume raw point cloud. Notice that our baseline does not include GEM and PAN. In our paper, we apply PyramNet to 3D point cloud object classification and semantic segmentation in 3D scene(See Fig. \ref{fig:3Dtasks}). We have experimented with standard datasets ModelNet40, ShapeNet and S3DID, and achieved good performance.

To summarize, the main contributions of our work are as follow:
\begin{itemize}
	\item We present a novel operator, PAN, which assigns each point some strong semantic features and retains the details of the geometry as much as possible;
	\item We also designed a new operator, GEM, to associate point cloud structure with graph. Then we exploit the covariance matrix to explore the relationship between points to enhance the local feature expression ability of the network;
	\item We embedded PAN and GEM modules into baseline to form a new point cloud processing structure, PyramNet, which effectively improves the performance of point cloud classification and semantics segmentation.
\end{itemize}

\section{Related Work}
\subsubsection{Voxel-based Methods}
3D data has a variety of expressions, such as voxels, meshes, and so on. Voxelization is a method of converting unstructured geometric data into a 3D mesh. Volumetric CNNs~\cite{Voxel_1} are the innovators applying CNN on 3D voxel. However, these voxel-based methods are often wasteful. Furthermore, these methods limit the resolution. Kdtree~\cite{kdtree} and Octree~\cite{octree} subdivide the spatial structure, but there are still resolution constraints.

\subsubsection{Multiviews-based Methods}
Point cloud are projected to 2D plane at small different angles. Networks based on multiview take advantage of different input including 2D rendered images and point cloud~\cite{miltuview_1,PvNet}. 2D projections may result in loss of surface information due to self-occlusion. And view point selection is usually achieved by heuristic, which is not necessarily optimal for a given task. These methods are also usually very computationally intensive.

\subsubsection{Graph-based Methods}
The Graph Convolutional Neural Network(GCNN) is well suited to process irregular data structures like point cloud.

One of the methods is that the convolution on graphs is defined in the spectral domain~\cite{about graph_3,about graph_6,about graph_5}. However, these methods need to calculate a large number of parameters. It is improved by polynomial or rational spectral filters~\cite{about graph_1,about grapg_2}.

Another graph-based approach is to implement convolution on each node and its neighbors, such as EdgeConv~\cite{DGCNN}. Compared with the spectral methods, its main advantage is that it is more consistent with the characteristics of data distribution. EdgeConv extracts edge features through the relationship between the central point and the neighbor points by constructing a graph.

\section{Our Approach}
We propose a novel deep learning network architecture, PyramNet, which can directly consume point cloud as input without other auxiliary input information, such as projections, pictures, etc. Consider a \textit{F}-dimensional point cloud with \textit{n} points $P=\{{{P}_{i}}|i=1,...,n\}\subseteq {{R}^{F}}$, where point $P_i$ is a set of vectors, including its coordinates $(x, y, z)$, and some additional features, such as RGB information, normal vector etc. We apply PyramNet in two different tasks: 3D point cloud object classification and semantic segmentation in 3D scenes.

Section \ref{subsec:PyramNet Arthitecturet} introduces the basic architecture of PyramNet. Section \ref{subsec:Graph Embedding Module} introduces a novel operator, Graph Embedding Module(GEM), which can capture local features of point cloud. Section \ref{subsec:Pyramid Attention Network} introduces another operator, Pyramid Attention Network(PAN), which can assign strong semantic features to each point. Section \ref{subsec:Network Hyperparameter} introduces the hyperparameters of PyramNet.

\subsection{PyramNet Arthitecture}
\label{subsec:PyramNet Arthitecturet}
\begin{figure}
	\includegraphics[width=\textwidth]{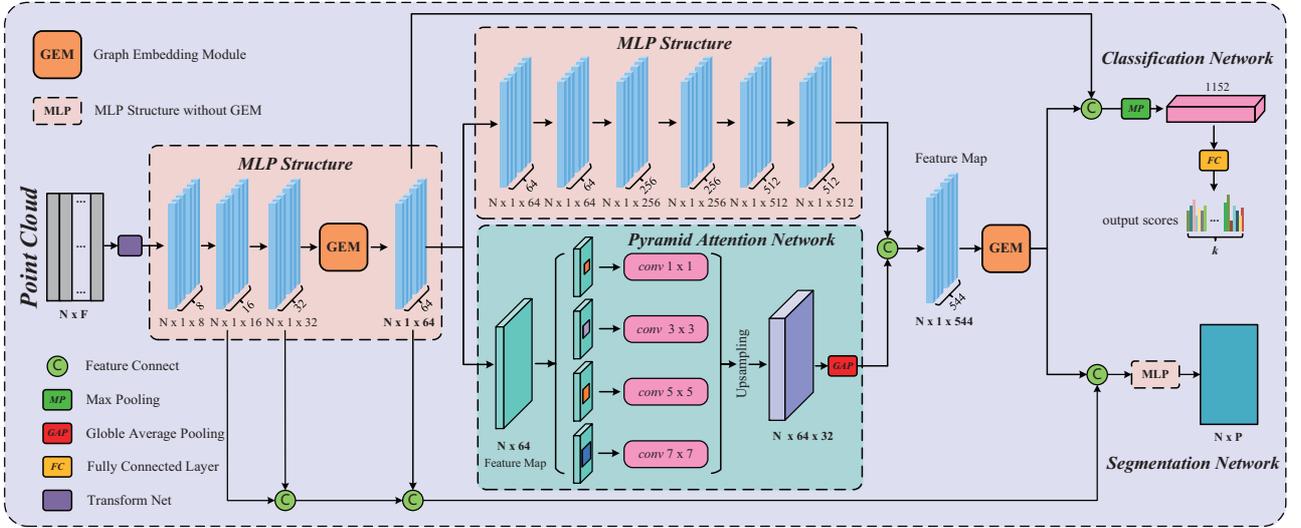}
	\caption{\textbf{PyramNet Arthitecture}. The PyramNet takes raw point cloud. It has two branches: classification branch and semantic segmentation branch. MLP structure, Graph Embeddeing Module and Pyramid Attention Network are the three main modules in PyramNet.} 
	\label{fig:PyramNet}
\end{figure}

In this section, we introduce our simple and effective network structure, PyramNet(See Fig. \ref{fig:PyramNet}).

PyramNet[\ref{fig:PyramNet}] has three critical parts. The \textit{MLP structure} is used for feature propagation. The \textit{Graph Embedding Module}(GEM) can better capture local geometric features between points. The\textit{ Pyramid Attention Network}(PAN) combines features of different resolutions and different semantic strengths, especially for semantic segmentation tasks in 3D scene.

We briefly describe our network structure. The input of the network is point cloud as N x F arrays, where N is the number of points in point cloud, and each point has \textit{F}-dimensional characteristics. The feature map(N x 1 x 32) is fed into the first GEM via multi layer perceptron. Then the feature map of N x 1 x 64 is spliced into N x 64. Next, the first parallel network including MLP structure and PAN starts working.

The top branch(MLP) outputs a feature map(N x 1 x 512). And the bottom branch outputs a feature map(N x 64 x 32) through PAN. After Global Average pooling(GAP), the feature map(N x 1 x 32) connects with the output of the top branch. Then the local feature is strengthened by the second GEM module. It is divided into two branches. Each branch connects a different number of shortcuts from the first MLP structure. \textit{Classification Network}(top branch) outputs the probability of each object through the max pooling layer and the fully connected layer. After MLP structure without GEM, the segmentation network outputs a N x P score table, P is the number of categories or classes.

\subsection{Graph Embedding Module}
\label{subsec:Graph Embedding Module}
As shown in Fig. \ref{fig:GEM}, GEM stands for Graph Embedding Module. Since point cloud is an irregularly distributed data structure, its main features are distributed in 3D space. Let's not consider the pyramid structure first. The role of MLP in Fig. \ref{fig:PyramNet} is to continuously select features on the properties of each point. In MLP structure, the size of the feature map is N x 1 x C. C represents the number of feature channels. N is constant, only C is changing. Hence, it is reasonable that the properties of each point are always changing during feature propagation. 

Assuming that each point has F attributes, that is, $A_{P}=\left\{A_{P_{i}} | i=1,2, \ldots, N\right\}$, $ A_{P_{i}}=\left\{A_{P_{i}}^{j} | j=1,2, \ldots, F\right\}$. $ A_{P} $ represents a set of attribute values of N points in the point cloud, and $ A_{P_{i}} $ represents a set of attribute values of each point. Except describing the relationship between points by using the characteristics of shared parameters of CNN, we also considere exploiting the geometry of point cloud to explore.

\subsubsection{Directed Acyclic Graph}
\begin{figure}
	\includegraphics[width=\textwidth]{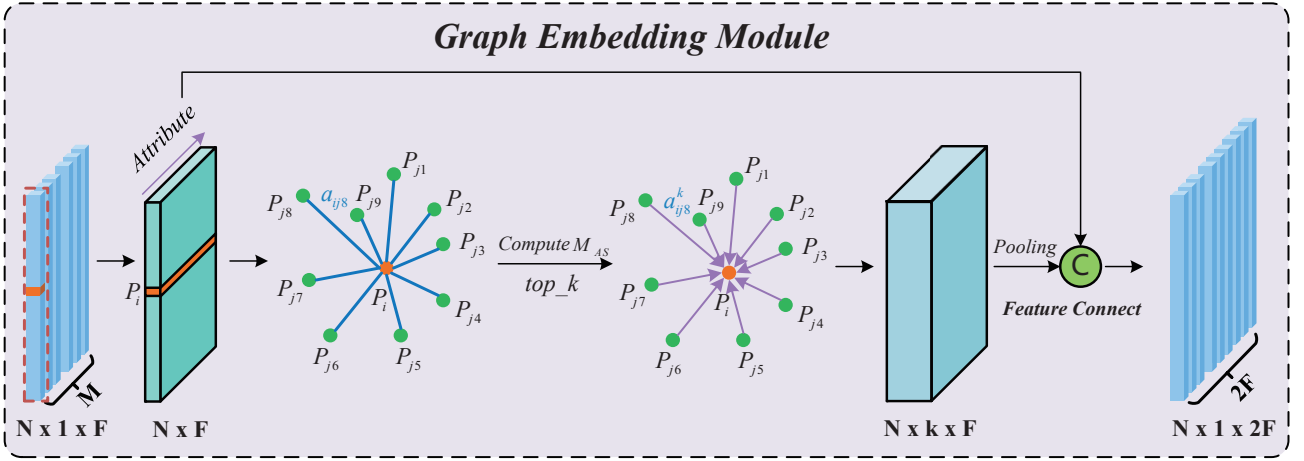}
	\caption{\textbf{Graph Embedding Module}. $ M_AS $ is the adjacency similarity matrix. $ a_{ij8}^{k} $ means the top-k attributes between Point $ P_i $ and $ P_j8 $. The input of GEM is N x 1 F feature map, while the output of GEM is N x 1 x 2F.}
	\label{fig:GEM}
\end{figure}

Inspired by the idea of graph convolutional Neural Network, we try to use the graph to describe the relationship between attributes $ A_{P_{i}} $ in point cloud. The purpose is to capture the local geometric features of point cloud. As shown in Fig. \ref{fig:GEM}, the input is N x 1 x F feature map, from which we can construct a directed acyclic graph. Graph $G=(V, E)$ represents the local structure of point cloud, where $V=N \subseteq R^{F}$ is a set of points in point cloud, and $E \subseteq V \times V$ is a set of edges between points. The attribute value of each point in the point set is $ A_{P_{i}} $. We define an adjacency similarity matrix to describe edge sets.

\subsubsection{Adjacency Similarity Matrix}
We project the features of each point into high dimensional space. Afterwards, we use a covariance matrix instead of Euclidean distance to describe the feature between points. That feature is described by relevance. Points with the same label have a greater correlation, and vice versa.

Assuming that there are N points in point cloud, the set of attribute values of each point is $A_{P_{i}}=\left\{A_{P_{i}}^{j} | j=1,2, \ldots, F\right\}$. Initially, the mean set $\mu$ of each point's attribute value is:
\begin{equation}
	\mu=\left\{\mu_{A}^{i} | i=1,2, \ldots, N\right\}
\end{equation}

where $\mu_{A}^{i}=E\left(A^{i}\right)$, $A^{i}$ is a collection of each point's attribute. Then we construct the covariance matrix $S \subseteq R^{N \times N}$. The term (i, j) of S is:
\begin{equation}
	S_{i j}={conv}\left(P_{i}, P_{j}\right)=E\left[\left(P_{i}-\mu_{A}^{i}\right)\left(P_{j}-\mu_{A}^{j}\right)\right]
\end{equation}
According to covariance matrix, for any point $ P_{i} $, retain the top k term which is the strongest correlation with the remaining N-1 points as the new attribute of $ P_{i} $. Therefore we get the \textbf{Adjacent Similarity Matrix} $M_{A S} \subseteq R^{N \times k}$.

Through the calculation of $M_{A S}$, the correlation between points is strengthened, as a result, the network becomes thicker. The feature map of N x 1 x F becomes N x k x F. Then, it goes through the global average pooling layer and is connected with the original input of GEM. The output of GEM becomes N x 2F. The information at each point becomes richer. Because the input channels of GEM differ greatly in the order of magnitude, we believe that the k is not fixed. After many experiments, we found that when $k=\left\lceil\frac{F}{4}\right\rceil$, the experiment works well(More details in section \ref{subsec:top-k}). Through experiments, we found that GEM played an important role in classification task and semantic segmentation.

\subsection{Pyramid Attention Network}
\label{subsec:Pyramid Attention Network}
The pyramid network is widely used in the semantic segmentation task in 2D world, and has achieved a great performance. We consider introducing this method into the 3D deep learning on point cloud in a reasonable way. Except for the first layer of the network and the Pyramid Attention Network, all other parts of the network use a 1 x 1 convolution kernel. 

The Pyramid Attention Network uses four different sizes of convolution kernels to downsample the feature map(See Fig. \ref{fig:PyramNet}). The reason why different convolution kernels are used is so that the features of adjacent points with different semantic intensities can be fused. It ensures that each layer is assigned some strong semantic features. The simple bilinear interpolation is used as upsampling to recover the details of the origin feature map.

The PAN can increase the receptive field and classify more efficiently, which further improves the accuracy of local feature extraction.

\subsection{Network Hyperparameter}
\label{subsec:Network Hyperparameter}
At the time of training, the classical cross entropy loss is used to supervise the classification branch and the semantic segmentation branch. We trained the network for 300 epochs on a single NVIDIA GTX 1080 GPU using Tensorflow1.6 with ADAM optimizer, initial learning rate 0.001, batch size 32, momentum 0.9. The decay rate for batch normalization starts with 0.5 and is gradually increased to 0.999. All layers include ReLU and batch normalization except for the last layer. In training, we use dropout with keep ratio 0.65 on the last fully connected layer in our classification architecture. Notice that dropout is not used in the semantic segmentation network. In testing, dropout is not used.

\section{Experiments}
\label{sec:experiment}
In this section, to verify the effectiveness of our model for processing point cloud, we evaluate our model on the ModelNet40~\cite{ModelNet}, ShapeNet~\cite{ShapeNet} and S3DIS~\cite{S3DIS}.

\subsection{3D Object Classification}

\subsubsection{Implementation Details}
The network architecture of the classification task is shown in the top branch in Fig. \ref{fig:PyramNet}. The input is raw point cloud as  N x F arrays. N is 1024 points uniformly sampled from the mesh faces and uniformly normalized to the unit sphere. We take F = 3, that is, we take the 3D coordinates($ x_i, y_i, z_i $) as the origin attribute of each point.

\subsubsection{Result}
We evaluate the performance of the 3D Object Classification on ModelNet40~\cite{ModelNet,PointNet}. Table \ref{tab:Classification} shows the results of our model in the classification task. The methods listed in table \ref{tab:Classification} have one thing in common. The input is only raw point cloud with 3D coordinates($ x_i, y_i, z_i $). 

Our baseline does not include GEM and PAN modules. As the table \ref{tab:Classification} shows, our baseline is slightly better(0.4\%) than PointNet~\cite{PointNet}. When the baseline is added to PAN, the accuracy is slightly better than the baseline. And when baseline is added to GEM, Our model achieves similar performance with PointNet++~\cite{PointNet++}. Finally, baseline + GEM + PAN achieves 91.5\% accuracy on ModelNet~\cite{ModelNet}, which demonstrates the validity of PyramNet.

\begin{table}
	\caption{\textbf{Classification Results on ModelNet40.}}
	\centering
	\label{tab:Classification}
	\setlength{\tabcolsep}{8mm}
	{
		\begin{tabular}{ccc}
			\toprule
			methods & \tabincell{c}{Accuracy \\ Avg Class} & \tabincell{c}{Accuracy \\Overall} \\
			\midrule
			PointNet~\cite{PointNet} & 86.2 & 89.2 \\
			PointNet++~\cite{PointNet++} & - & 90.7 \\
			OctNet~\cite{octree} & 83.8 & 86.5 \\
			Kd-Net~\cite{kdtree} & 88.5 & 91.8 \\
			EdgeConv~\cite{DGCNN} & \textbf{90.2} & \textbf{92.2} \\
			\midrule
			Baseline & 86.7 & 89.6 \\
			Baseline+PAN & 87.0 & 89.9 \\
			Baseline+GEM & 87.6 & 90.6 \\
			Ours PyramNet & 88.3 & 91.5 \\
			\bottomrule
		\end{tabular}
	}
\end{table}

\subsection{3D Semantic Segmentation}
\label{subsec:semseg}
There are two types of point cloud semantic segmentation in 3D scenes. Part segmentation is to assign a predefined part category label(e.g. laptop screen, airplane aerofoil) to each point. 3D scene semantic segmentation is to assign a semantic objetc class(e.g. chair, laptop in a room) to each point for a given 3D object model.

\begin{table}
	\caption{\textbf{The Result of Part Segmentation on ShapeNet}. The evaluation indicator is mean IoU(\%) on points. We choose 11 of these categories(16 in total) to show the results.}
	\centering
	\label{tab2}
	{
		\begin{tabular}{c|c|ccccccccccc}
			\toprule
			& mIoU & aero & bag & cap & car & chair & \tabincell{c}{ear \\ phone} & guitar & knife & lamp & laptop & motor  \\
			\midrule
			\tabincell{c}{Shapes \\ Numbers} &  & 2690 & 76 & 55 & 898 & 3758 & 69 & 787 & 392 & 1547 & 451 & 202  \\
			\midrule
			PointNet~\cite{PointNet} & 83.7 & 83.4 & 78.7 & 82.5 & 74.9 & 89.6 & 73.0 & 91.5 & 85.9 & 80.8 & 95.3 & 65.2  \\
			PointNet++~\cite{PointNet++} & \textbf{85.1} & 82.4 & 79.0 & \textbf{87.7} & 77.3 & 90.8 & 71.8 & 91.0 & 85.9 & \textbf{83.7} & 95.3 & \textbf{71.6}  \\
			Kd-Net~\cite{kdtree} & 82.3 & 80.1 & 74.6 & 74.3 & 70.3 & 88.6 & 73.5 & 90.2 & 87.2 & 81.0 & 94.9 & 57.4 \\
			EdgeConv~\cite{DGCNN} & \textbf{85.1} & 84.2 & \textbf{83.7} & 84.4 & 77.1 & \textbf{90.9} & \textbf{78.5} & 91.5 & \textbf{87.3} & 82.9 & 96.0 & 67.8 \\
			\midrule
			\tabincell{c}{Ours \\ PyramNet} & 83.9 & \textbf{84.4} & 81.3 & 80.4 & \textbf{77.5} & 94.5 & 67.9 & \textbf{91.8} & 86.4 & 70.6 & \textbf{96.8} & 66.3 \\
			\bottomrule
		\end{tabular}
	}
\end{table}

\subsubsection{Implementation Details}
As shown in the bottom branch of Fig. \ref{fig:PyramNet}, the size of the input is still N x F. In part segmentation task, N is 2048, F is 3, while N is 4096, F is 9 in semantic segmentation in 3D scene task. All points are randomly and uniformly sampled. During training, we use the same tricks as the classification task to augment point cloud.

The segmentation is a simple extension of the classification network. But the difference is that the second GEM is connected with several shortcuts from the first MLP structure. At last, three shared fully connected layers (512,256,P) are used to transform the pointwise features. Thence, the semantic segmentation network outputs a N x P probability map. P is 50 in part segmentation and P is 13 in semantic segmentation in 3D scene.

\subsubsection{Result}
\begin{figure}
	\centering
	\subfigure[GT]{
		\includegraphics[width=0.22\textwidth]{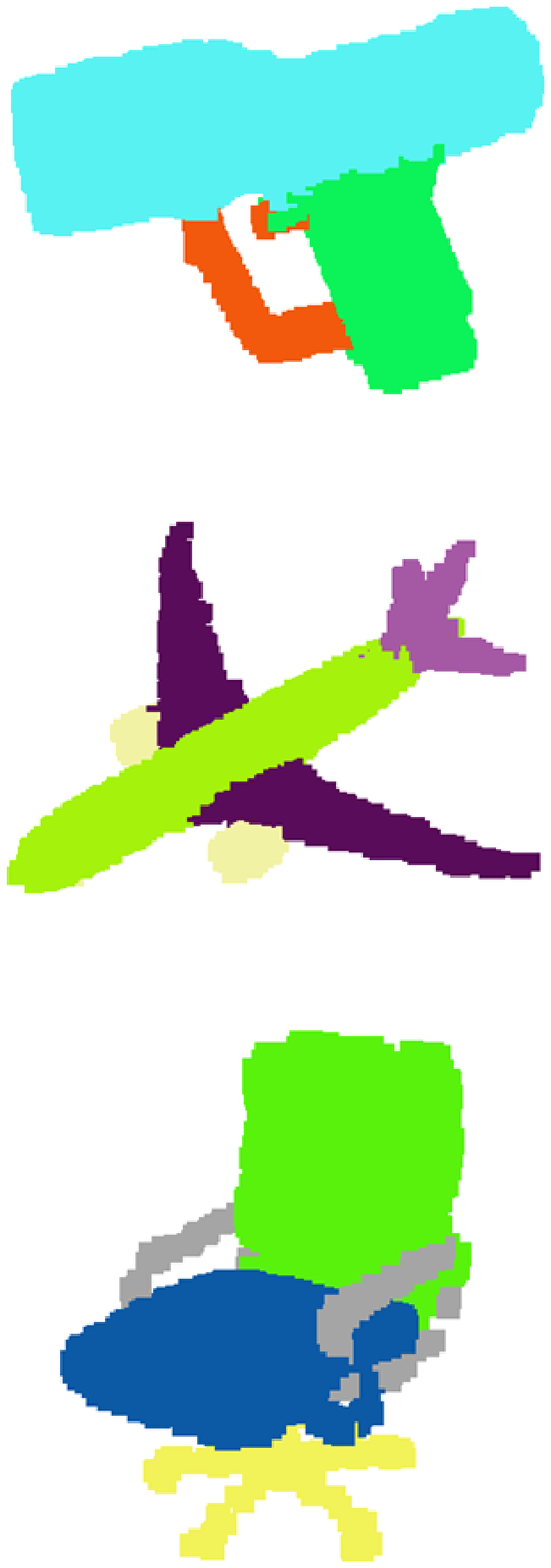}
		\label{fig:Part_GT}
	}
	\subfigure[Baseline]{
		\includegraphics[width=0.22\textwidth]{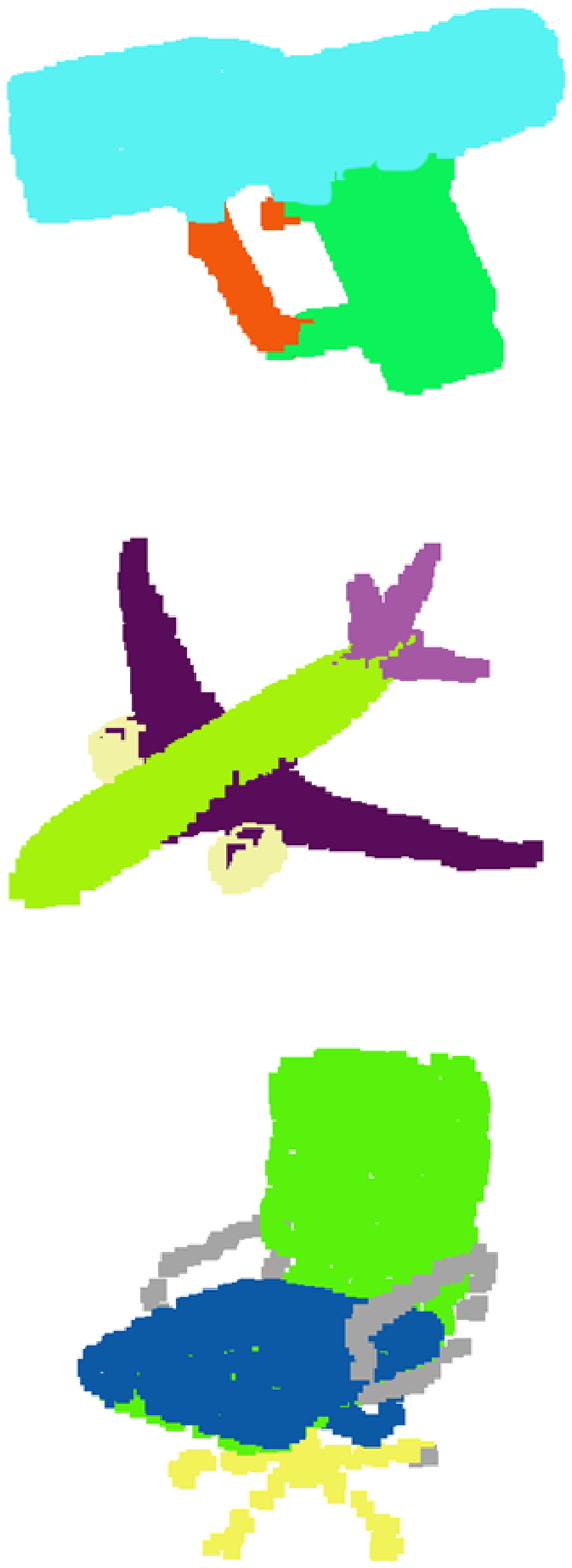}
		\label{fig:Part_Baseline}
	}
	\subfigure[PyramNet]{
		\includegraphics[width=0.22\textwidth]{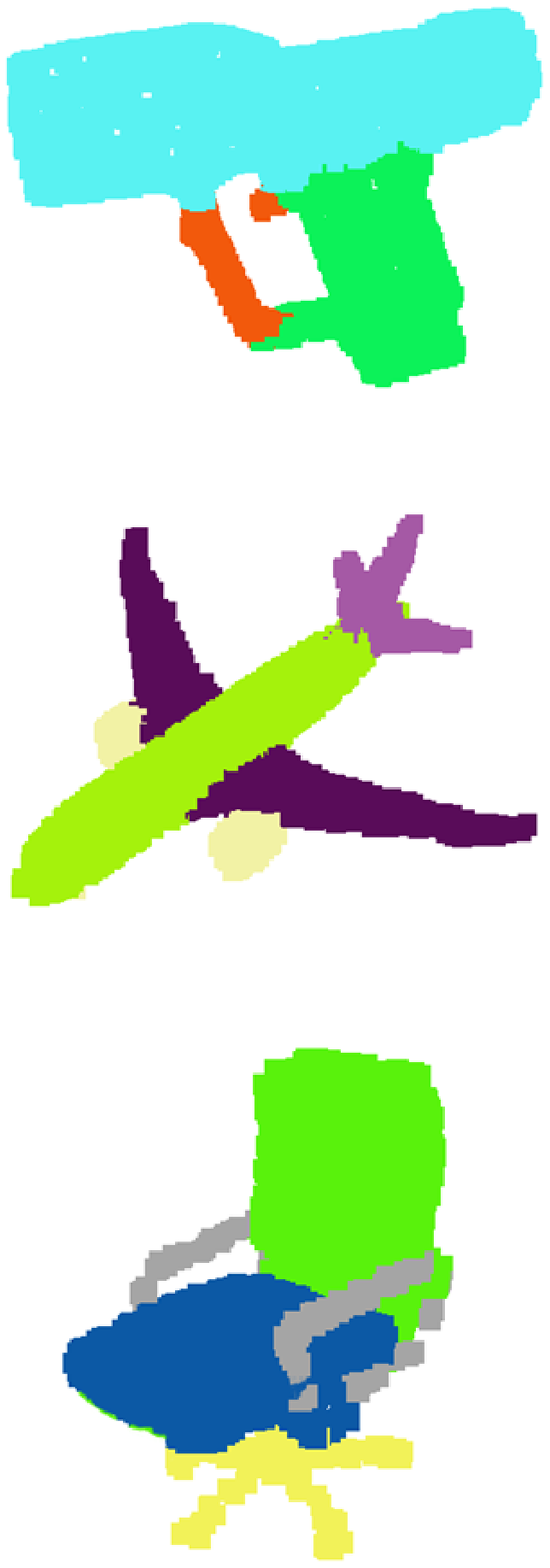}
		\label{fig:Part_PyramNet}
	}
	\subfigure[Difference]{
		\includegraphics[width=0.22\textwidth]{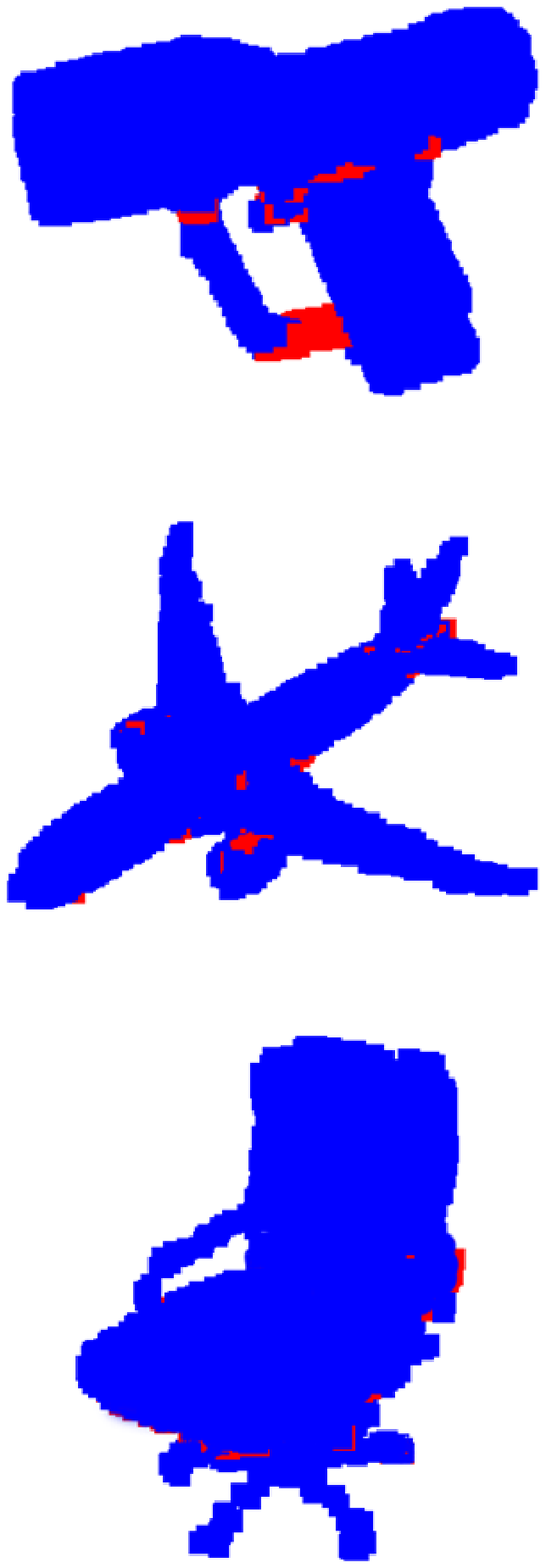}
		\label{fig:Part_Difference}
	}
	
	\caption{\textbf{Part Segmentation Results on ShapeNet}. From left to right: Ground True, Baseline, PyramNet, Difference. The red color in the last column \ref{fig:Part_Difference} indicates the difference area between PyramNet \ref{fig:Part_PyramNet} and Ground True \ref{fig:Part_GT}.} 
	\label{fig:ParSeg}
\end{figure}

\paragraph{\textbf{Part Segmentation}}
We evaluated the performance of the part segmentation on ShapeNet~\cite{ShapeNet}. We compare our model with PointNet, PointNet++, Kd-Net and EdgeConv. In table 2, we list all categories and mIOU scores. The mIOU of our model is 83.9\%. It's slghtly better than PointNet and Kd-Net. Our model achieves the best results in some categories. We save the results in obj format and visualize the part segmentation(See Fig. \ref{fig:ParSeg}) using meshlab.

\begin{table}
	\caption{Result of S3DIS}
	\centering
	\label{tab:SemSeg}
	\setlength{\tabcolsep}{4mm}
	{
		\begin{tabular}{ccc}
			\toprule
			& mIoU & Accuracy Overall \\
			\midrule
			PointNet~\cite{PointNet} & 47.7 & 78.6 \\
			EdgeConv~\cite{DGCNN} & \textbf{56.1} & 84.1 \\
			\midrule
			Ours(PyramNet) & 55.6 & \textbf{85.6} \\
			\bottomrule
		\end{tabular}
	}
\end{table}

\paragraph{\textbf{3D scene semantic segmentation}}
We evaluated the performance of the semantic segmentation in 3D scene on S3DIS~\cite{S3DIS}. We predict the sementic object class for each point. We compare our model with PointNet~\cite{PointNet} and EdgeConv~\cite{DGCNN} in table \ref{tab:SemSeg}. Our model has a significant improvement over PointNet. We observe a 1.5\% accuracy improvement compared with EdgeConv.In terms of mIoU, it is similar to EdgeConv.

As shown in Fig. \ref{fig:SemSeg}, we choose the offices 1, 2, 5, 21, and 36 in area6 in the S3DIS for visualization. From the visualization results, we can find that PyramNet can effectively segment objects of different semantic categories. Furthermore, with the help of GEM and PAN, PyramNet outperforms the baseline significantly.


\subsection{Ablation Study}

\subsubsection{Effectiveness of GEM and PAN}
\begin{figure}
	\centering
	\subfigure[Real Scene]{
		\includegraphics[width=0.22\textwidth]{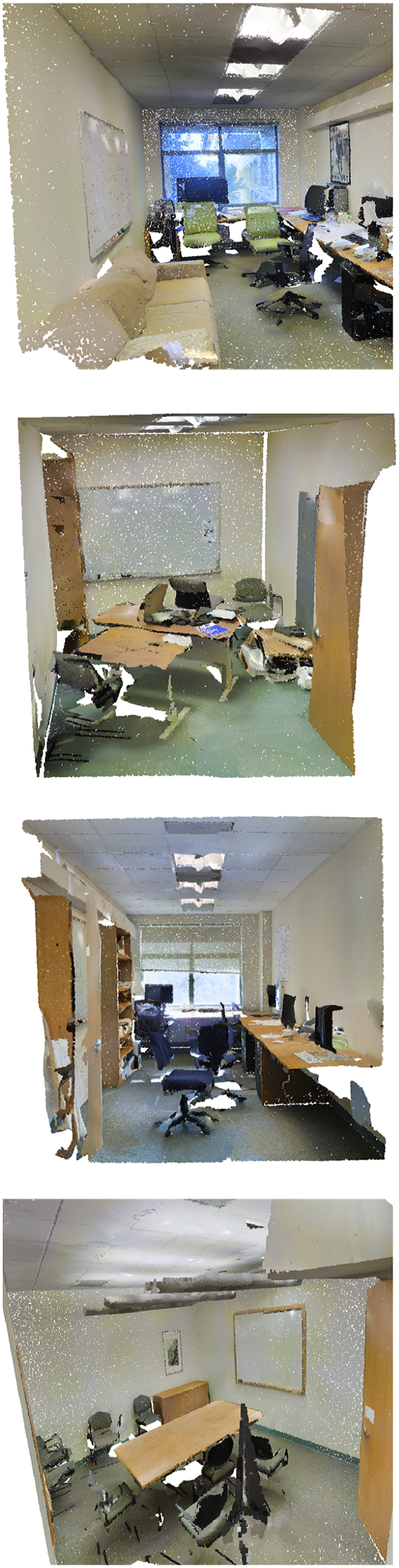}
		\label{fig:Seg_Real Scene}
	}
	\subfigure[Ground True]{
		\includegraphics[width=0.22\textwidth]{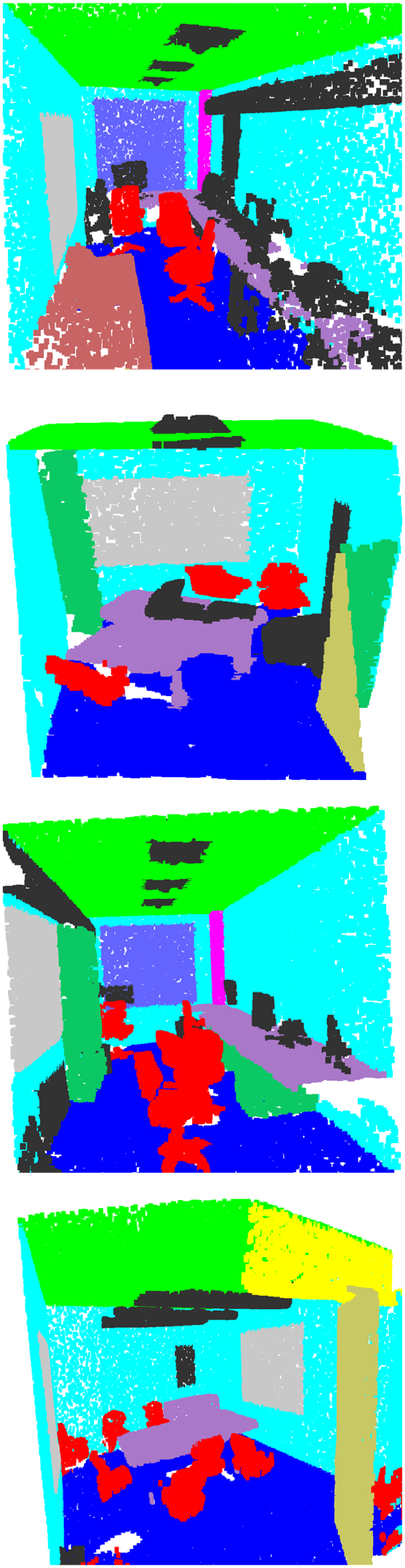}
		\label{fig:Seg_GT}
	}
	\subfigure[Baseline]{
		\includegraphics[width=0.22\textwidth]{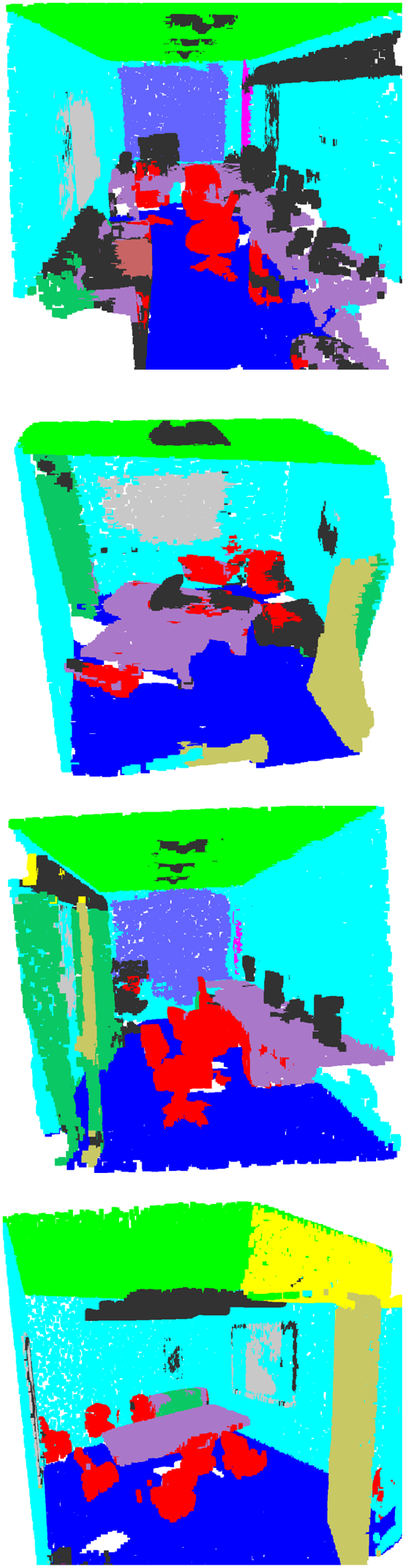}
		\label{fig:Seg_Baseline}
	}
	\subfigure[PyramNet]{
		\includegraphics[width=0.22\textwidth]{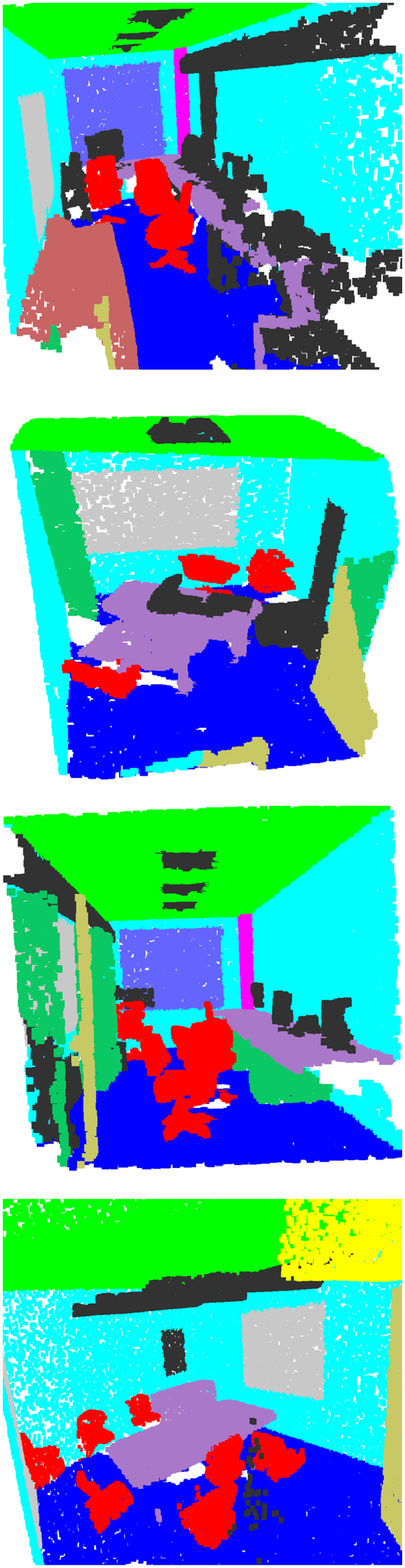}
		\label{fig:Seg_PyramNet}
	}
	
	\caption{\textbf{3D Scene Semantic Segmentation Results on S3DIS}. From left to right: Real Scene \ref{fig:Seg_Real Scene}, Ground True \ref{fig:Seg_GT}, Baseline \ref{fig:Seg_Baseline}, PyramNet \ref{fig:Seg_PyramNet}.} 
	\label{fig:SemSeg}
\end{figure}

In Fig. \ref{fig:ParSeg} and Fig. \ref{fig:SemSeg}, we visualize the results of Baseline and PyraNet. For example, in 3D scene semantic segmentation task, the baseline is prone to semantic label migration, which indicates that a powerful semantic label may encroach on another weakly expressive semantic label. While GEM and PAN can effectively separate different semantic classes or different object part labels.

\subsubsection{The top-k in GEM}
\label{subsec:top-k}
In section \ref{subsec:Graph Embedding Module}, we introduce the k value selection method in the GEM module. More specifically, the input size of the first GEM is N x 1 x 32, while the input size of the second GEM is N x 1 x 544. These two input channels differ by an order of magnitude. We take k = 20, 30 and $\left\lceil\frac{F}{4}\right\rceil$ to do some experiments(Table \ref{tab4}), where F is channels of the input of GEM. After many experiments, we found that it is reasonable when $k=\left\lceil\frac{F}{4}\right\rceil$, PyramNet achieves 91.5\% accuracy.

\begin{table}
	\caption{The impact of k in GEM on PyramNet.}
	\centering
	\label{tab4}
	\setlength{\tabcolsep}{4mm}
	{
		\begin{tabular}{cc}
			\toprule
			k & Accuracy Overall \\
			\midrule
			20 & 88.6 \\
			30 & 90.1 \\
			\midrule
			$\left\lceil\frac{F}{4}\right\rceil$ & \textbf{91.5} \\
			\bottomrule
		\end{tabular}
	}
\end{table}

\section{Conclusion}
In this work, we propose a novel end-to-end deep learning framework, PyramNet, to directly consume point cloud. We propose two different operators, GEM and PAN, to learn point cloud in order to perform point cloud classification, part segmentation, and 3D scene semantic segmentation tasks. GEM and PAN can successfully learn the spatial local geometric features of point cloud. GEM and PAN also demonstrate that we should not only consider the rationality of the network, but also the geometric characteristics of 3D point cloud when we apply deep learning on it. Furthermore, for the irregular data such as 3D point clouds, the graph in the GEM module is an extensible research idea.


%
%
%
%
%

\end{document}